# Integrating a Heterogeneous Graph with Entity-aware Self-attention using Relative Position Labels for Reading Comprehension Model

**Shima Foolad**
Department of Electrical & Computer Engineering
Semnan University
Semnan, Iran
sh.foolad@semnan.ac.ir

**Kourosh Kiani***
Department of Electrical & Computer Engineering
Semnan University
Semnan, Iran
kourosh.kiani@semnan.ac.ir

July 21, 2023

## Abstract

Despite the significant progress made by transformer models in machine reading comprehension tasks, they still fall short in handling complex reasoning tasks due to the absence of explicit knowledge in the input sequence. To address this limitation, many recent works have proposed injecting external knowledge into the model. However, selecting relevant external knowledge, ensuring its availability, and requiring additional processing steps remain challenging. In this paper, we introduce a novel attention pattern that integrates reasoning knowledge derived from a heterogeneous graph into the transformer architecture without relying on external knowledge. The proposed attention pattern comprises three key elements: global-local attention for word tokens, graph attention for entity tokens that exhibit strong attention towards tokens connected in the graph as opposed to those unconnected, and the consideration of the type of relationship between each entity token and word token. This results in optimized attention between the two if a relationship exists. The pattern is coupled with special relative position labels, allowing it to integrate with LUKE's entity-aware self-attention mechanism. The experimental findings corroborate that our model outperforms both the cutting-edge LUKE-Graph and the baseline LUKE model across two distinct datasets: ReCoRD, emphasizing commonsense reasoning, and WikiHop, focusing on multi-hop reasoning challenges.

***Keywords*** global-local attention. graph-enhanced self-attention. transformer-based model. reasoning knowledge. relative position encoding · cloze-style machine reading comprehension.

## 1 Introduction

Recent successes on a variety of Natural Language Processing (NLP) tasks, including Question Answering (QA) and Machine Reading Comprehension (MRC), have been achieved using transformer-based models, such as BERT [1] or other variations. The main innovation in transformers is the addition of a self-attention mechanism that can evaluate each token of the input sequence simultaneously. Thanks to this parallelism, transformers may train NLP models on datasets of unprecedented size, allowing contemporary hardware accelerators like GPUs/TPUs to fully leverage their capabilities. This ability enables the models to be pretrained on huge general-purpose corpora, which then allows the knowledge to be transferred to downstream tasks like QA and MRC.

Despite the advanced outcomes achieved by most transformer-based models, their transformer architecture presents two notable limitations: (1) Inclusion of Reasoning Information: the challenge lies in enriching a specific pre-training objective with reasoning information, and (2) Constraints of Attention Allocation between Tokens: determining how to implement the relationship type of each pair of tokens (such as an entity token with its mention, a query token with another token) in the attention assigned to each other is another challenge.

---
*Corresponding author

The original transformer model [2] has the ability to link tokens (words) to each other via the self-attention mechanism. However, managing these relationships within datasets that necessitate strong commonsense reasoning, such as ReCoRD, proves to be a difficult task, given that the model typically breaks down most entities into multiple tokens. In order to address this challenge, several recent studies [3–5] have explored the use of external knowledge to enhance the reasoning capabilities of models. On the other hand, some studies [4,6–9] have introduced entities as distinct tokens within the model, allowing for direct comprehension of the relationships between entities. In some of them[4,9], each entity is allocated a fixed embedding vector that stores information about the entity in a knowledge base (KB). Whilst these models are capable of capturing the rich information present in the KB, they are limited in their capacity to represent entities that do not exist within the KB. Moreover, they necessitate the selection of the optimal subgraph and the most pertinent entity in KGs, particularly for ambiguous ones. By contrast, other researches [6,8] have proposed contextual representations of entities that are trained using unsupervised pre-training tasks based on language modeling. Most of the research overlooks the inclusion of prior knowledge about linking mentions of different entities in documents and the intuitive relationships between them. The relationships contain abundant reasoning information in documents.

Conversely, the original transformer model simply connects the tokens together, without considering their relationship type in their attention to each other. Various models of the transformer [8,10–13] have been suggested for scaling up input length and reducing the memory and computational requirements in the full self-attention mechanism. They have adapted the transformer architecture to employ sparse attention, which restricts the ability of tokens to attend to one another, by applying specialized attention patterns. As such, they identify some input tokens as local, attending to nearby tokens, and some important ones as global, attending to all tokens (e.g., [CLS] token). The approach most closely related to this paper is the ETC model [8]. This model introduces extra global tokens that do not correspond to any of the input tokens. In addition, it has been proposed to manage structured inputs by amalgamating global-local attention with relative position labels [14].

In order to address the aforementioned limitations, we propose a Graph-Enhanced Self-Attention (GESA) approach without relying on external knowledge. First, we employ the LUKE [6] as the foundational pre-trained model, which considers both words and entities in a given document as distinct input tokens. Then, we augment the entity-aware self-attention mechanism of LUKE by dividing the attention matrix into four components: w2w, w2e, e2w, and e2e, depending on token types - word (w) or entity (e). For the w2w component, we utilize the global-local attention approach used in Longformer [10] by designating certain input tokens as local, to concentrate on nearby tokens, and some key ones ([CLS] and question tokens) as global, to attend all tokens. For the w2e and e2w components, we efficiently account for the relationship type between each entity token with word token (e.g., entity token with its word mention in the document, question tokens with missing entity token) in the attention given to each other. Based on the relationship type of each pair, we assign a unique relative position embedding and add it to the scaled dot product operation. Additionally, we transform all entity candidates in the document and their connections into a heterogeneous graph and integrate the graph information into the e2e part of the attention matrix. Our experimental results showcase GESA's superior performance compared to both the advanced LUKE-Graph [7] and the baseline LUKE model [6] on the ReCoRD dataset [15] (emphasizing commonsense reasoning), as well as the WikiHop dataset (known for its multi-hop reasoning challenges). In summary, our primary contributions are:

- We introduce a novel attention pattern that includes global-local attention for word tokens, graph attention for entity tokens, and attention between related entity and word tokens. This attention pattern seamlessly integrates commonsense representations into the fine-tuning phase without using any external knowledge.
- We incorporate this attention pattern into the entity-aware self-attention mechanism of LUKE using relative position encoding. This integration allows the self-attention mechanism to effectively emphasize reasoning information, thereby aiding in accurate decision-making.
- By incorporating the reasoning information from a heterogeneous graph into the self-attention mechanism, we allow the transformer layers to give more attention to tokens connected in the graph than those unconnected.
- By considering the relationship type between tokens, we modify the attention mechanism to limit their mutual attention, leading to enhanced attention efficiency when dealing with related tokens.
- We evaluate the effectiveness of our proposed method across datasets that demand two different types of reasoning—commonsense reasoning in the ReCoRD dataset and multi-hop reasoning in the WikiHop dataset. The outcomes showcase a notable performance improvement, indicating a 1.1% and 5.6% enhancement compared to the baseline LUKE model for the respective datasets.

## 2   Related Work

In this section, we review methods that improve the original transformer by scaling up input length, longer-term attention span, limiting the connection between tokens in self-attention, and reducing the memory consumption and computation time. The attention span of the original transformer is fixed and constrained. Therefore, no information can cross fixed-length segments. The Transformer-XL [16] extended the attention span over a longer period of time and across several segments by incorporating information from earlier hidden states. Moreover, to maintain a consistent flow of positional information across segments, it encoded a relative position rather than absolute position. It has been shown that relative position embeddings perform better in tasks requiring comprehension and the creation of natural language [14,16]. To reduce memory and computational requirements, Sparse Transformer [12] introduced factorized self-attention using predefined sparsity patterns such as local or stride attention, making it possible to train deeper networks on sequences of unprecedented length on current hardware.

The objective of research lines like Routing Transformer [17], Reformer [13], and Sinkhorn Transformer [18] is to acquire knowledge of sparsity patterns. In particular, Routing Transformer [17] implements a sparsity pattern learning method that is based on content similarity and employs k-means clustering. This approach allows it to generate attention queries and keys exclusively from the same cluster. On the other hand, the Reformer [13] model uses locality-sensitive hashing (LSH) [19] to incorporate attention mechanisms and to compute attention only for query and key vectors within the same hash buckets.

Another line of study [8,10,11] showed both attention types (local and global attention) are essential. The local attention is mainly utilized to construct contextual representations, while the global attention can build whole sequence representations for prediction. Longformer [10] introduced a windowed local-context self-attention along with global attention to a few pre-selected input tokens for learning task-specific representations, such as CLS token for classification and all question tokens for MRC. Extended transformer construction (ETC) model [8] defined some additional global content embeddings that do not correspond to any of the input tokens. The Big-Bird [11] also added random sparse attention patterns to global-local attention in the ETC structure. It enables the quick blending of information from different parts of the input sequence. Since the ETC model can directly encode a graph or hierarchical structure, we model its attention pattern for integrating our heterogeneous graph into a global-local attention. Moreover, we expand the vocabulary of relative position labels to efficiently consider the relation type of each pair of tokens in the amount of attention to each other. Comparing our attention pattern with other attentions is shown in Fig. 1. In our attention pattern, three forms of attention, namely the global-local attention used in Longformer, graph attention, and attention pattern employed in ETC, are combined. The attention matrix is partitioned into four segments, namely w2w, w2e, e2w, and e2e, based on the types of tokens, i.e., word (w) or entity (e), inspired by ETC. The global-local attention mechanism of Longformer is incorporated into the w2w segment, wherein some input tokens are treated as local to attend to nearby tokens, while certain crucial ones ([CLS] and question tokens) are deemed global and attend to all tokens. The relationship type between each entity token and word token is effectively considered in the w2e and e2w segments, enabling them to attend to each other more strongly if they are related. Moreover, all entity candidates in the document and their connections are transformed into a heterogeneous graph, and the graph information is integrated into the e2e portion of the attention matrix.

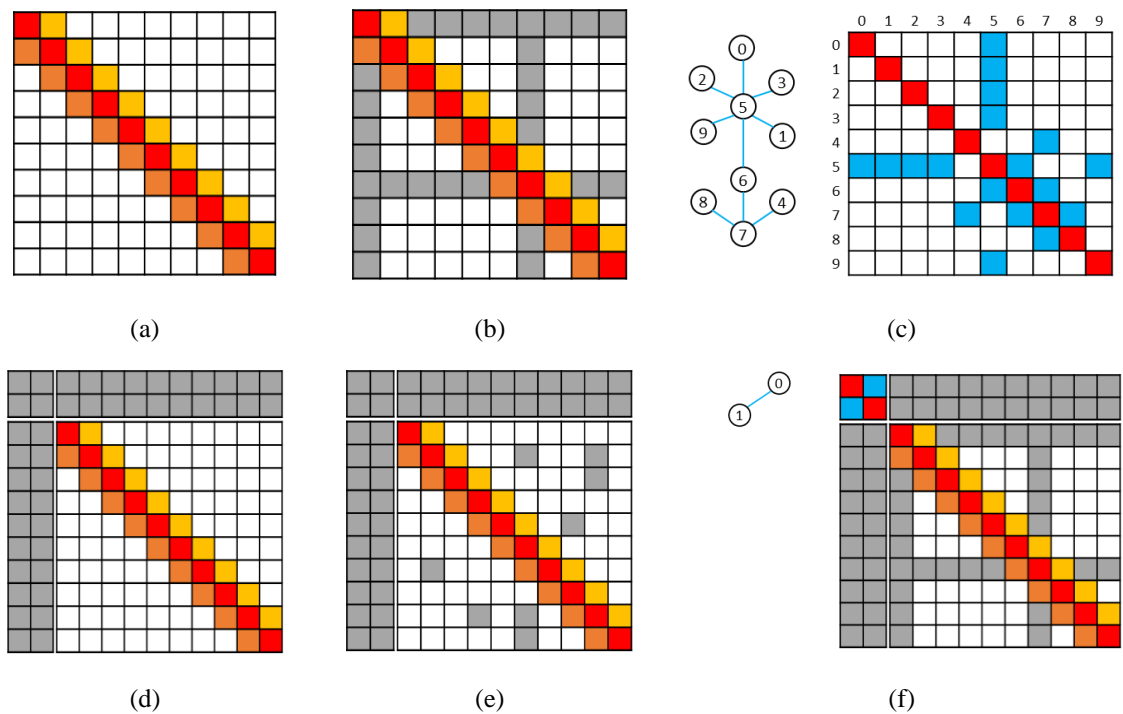

**Fig. 1.** Comparing different attention patterns with our model. (a) local attention. (b) global-local attention used in Longformer model. (c) graph attention. (d) global-local attention used in ETC. (e) global-local attention used in big-bird. (f) our attention

A different line of work [4–7,20–22] improved transformer-based models by injecting them with external knowledge. Li et al. [5] utilized pre-calculated embeddings from ConceptNet [23] as a form of external knowledge representation, which they incorporated into BERT through three different methods during the fine-tuning process. Additionally, they introduced a mask mechanism that allows for the token-level search of multi-hop relationships to filter external knowledge. Li et al. [24] introduced three simple yet effective techniques for injecting external commonsense knowledge into BERT. The method enables the models to utilize commonsense knowledge explicitly, enhancing their interpretability and ability to answer commonsense-related questions. Also, Meng et al. [21] proposed a confidence-based knowledge integration (CBKI) module, which determines the amount of knowledge to be integrated into the model based on confidence scores. However, external knowledge sources may not always be readily available and injecting them into the models requires additional processing steps. This can increase the complexity of the model, making it harder to train and deploy. ERNIE [4] and Know-BERT [20] learned static entity embeddings from a Knowledge Base (KB), while LUKE (Language Understanding with Knowledge-based Embeddings) applied a new pretraining task to learn entity representations. LUKE leverages entity information in its self-attention mechanism, allowing the model to attend to specific entities in the input text, thereby improving its performance on tasks that require an understanding of entity relationships. Besides, LUKE-Graph [7] combined the strengths of the LUKE with graph-based reasoning to capture intuitive relationships between entities. Integrating Gated Relational Graph Attention (Gated-RGAT) achieved a notable improvement. In this paper, we incorporate the knowledge from KB into our model using the LUKE pretrained task, and add the entity-aware self-attention mechanism to take the kind of the tokens (words or entities) into account while computing attention scores. Furthermore, we take advantage of the LUKE-Graph to capture the relationship importance between entities using a heterogeneous graph. however, we integrate the graph module with the self-attention mechanism instead of using a separate graph module from the transformer. This implies that a multi-step model will no longer be necessary, thereby resulting in a reduction in the model's execution time.

Recently, some Large Language Models or LLMs [25–27] have proposed with hundreds of billions of parameters that have achieved impressive results on various NLP tasks. DeBERTa (Decoding-enhanced BERT with disentangled attention) [26] introduces a disentangled attention mechanism and a decoding-enhanced training procedure to improve the efficiency and effectiveness of the BERT model. The disentangled attention mechanism in DeBERTa helps to reduce the impact of irrelevant information and increase the importance of relevant information in the attention mechanism. Moreover, a decoding-enhanced training procedure trains the model to generate the output sequence in a left-to-right order, rather than a random order. In contrast, T5 [25] is a text-to-text transformer that is pre-trained on a variety of tasks and can perform a wide range of natural language processing tasks by converting them into a text-to-text format, while PaLM (Pathways-augmented Language Model) [27] combines traditional transformer-based models with a novel pathways mechanism that allows for more efficient processing of long sequences. The pathways mechanism works by dividing the input sequence into smaller segments and processing each segment separately. One common disadvantage among T5, DeBERT, and PaLM models is their computational cost. All three models are large and complex, requiring significant computational resources to train and fine-tune. This can make it challenging to deploy these models on resource-constrained devices. While, the smaller number of parameters is one of our model key advantages, as it allows for more efficient training and inference.

## 3 Methodology

Fig. 2 displays the architecture of our model, which builds upon the pre-trained multi-layer transformer from the LUKE model [6]. Our model incorporates a number of significant modifications, the most notable of which is the construction of a heterogeneous graph that is linked to each token's designated relationship type as an attention pattern. This attention pattern is then integrated into the self-attention mechanism of our transformer using relative position encoding. To provide further context on our model, we present an outline below. Firstly, we partition the inputs into two distinct sequences, namely the word input for question and document tokens, including special characters such as [CLS] and [SEP], and the entity input for candidate answers or extracted entities originating from the document. Then, we compute a representation for each token in an embedding layer. Since our model uses relative position encodings in transformer layers, we exclude absolute position encodings in the embedding layer. Additionally, we build a heterogeneous graph based on relationships stemming from the entity input. The heterogeneous information regarding entities and the input sequence representations are imported to the transformer layers. In the multi-head attention part of the layers, an entity-aware self-attention mechanism of LUKE is modified by segregating the attention matrix into four parts: w2w, w2e, e2w, and e2e, depending on token types - word (w) or entity (e). For the w2w part, we use the global-local attention approach utilized in Longformer [10] by designating certain input tokens as local, to focus on nearby tokens, and some critical ones ([CLS] and question tokens) as global, to attend all tokens. For the w2e and e2w portions, we efficiently account for the relationship type between each entity token with word token, enabling them to attend to each other more strongly if they are related. Based on the relationship type of each pair, we assign a unique relative position embedding and add it to the scaled dot product operation. For example, one relative position label is assigned to link the entity tokens with the word tokens that belong to them, and a different label for those that do not.

Additionally, we integrate the heterogeneous graph information into the e2e part of the attention matrix, enabling it to attend more strongly to connected tokens in the graph than those unconnected. Finally, we compute a score for each candidate entity using a linear classifier in a score accumulation part and choose the candidate with the highest score as the final answer.

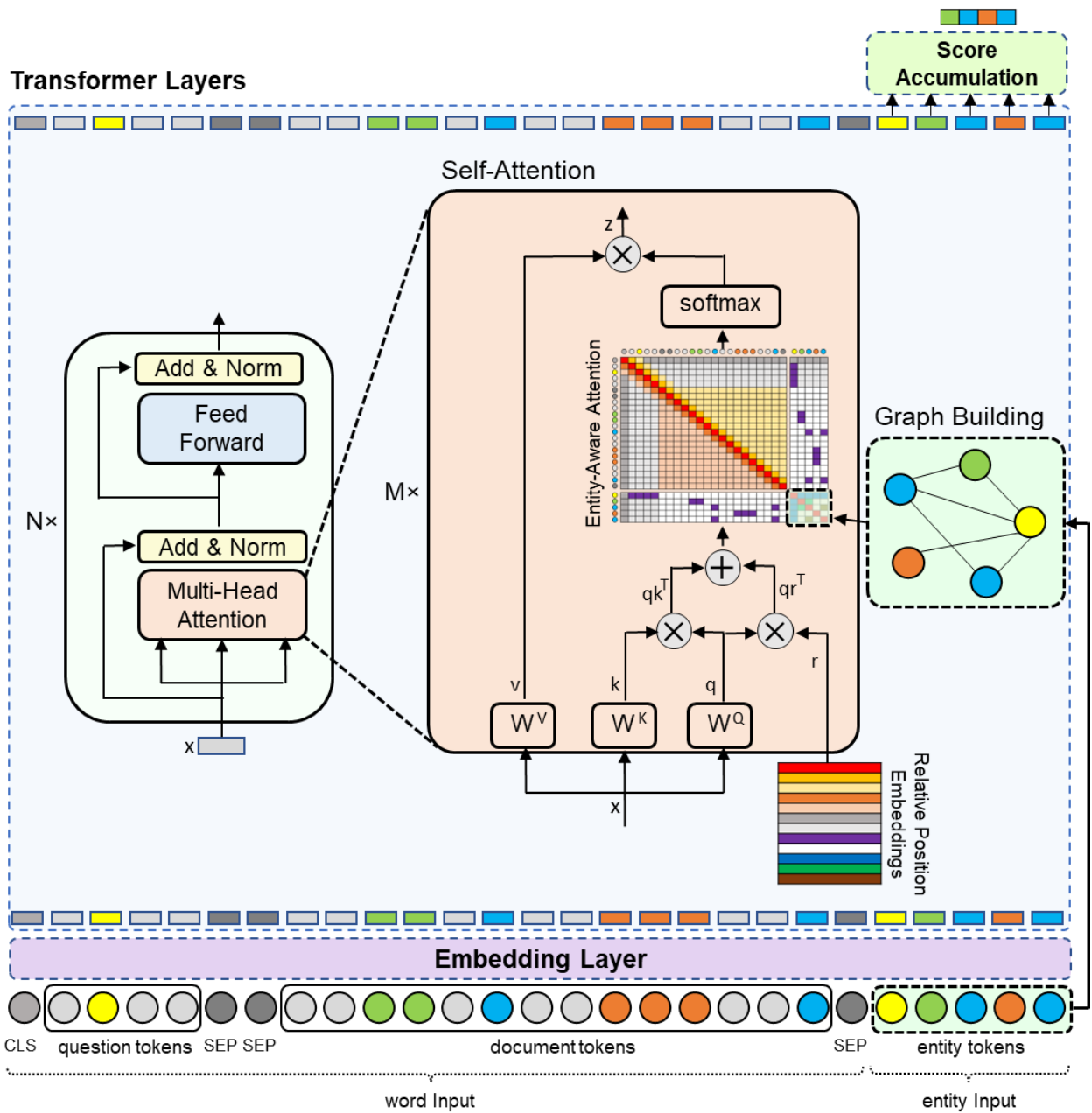

**Fig. 2.** The architecture of our method.

### 3.1 Embedding Layer

The Embedding layer utilizes the LUKE pre-trained language model to encode text to textual representations. For cloze-style reading comprehension task, it takes the following input sequence: {[CLS], Q, [SEP], [SEP], D, [SEP], E} where Q represents all question tokens with a [PLC] special token in the position of missing entity (placeholder), D refers to all document tokens along with two [ENT] special tokens for each entity within it as entity separators, and E denotes [MASK] tokens for missing entity and each candidate entity appearing in the document. In cases where entities are not explicitly specified within the dataset, such as the WikiHop dataset, we employ the entity extraction method outlined in [7]. We consider E as the entity input and others as the word input. Further, [CLS] and [SEP] special tokens are defined as a separator and a classification token, respectively.

In our modified transformer model, we begin by computing a representation for each token in an embedding layer. The input representation of a token is computed using the following two embeddings: token embedding and entity type embedding. A token embedding is a numerical representation of the corresponding token learned during pretraining the LUKE model. The entity type embedding defines the type of token, whether it is a word or an entity. Since our model uses relative position encodings in transformer layers, we omit absolute position encodings in the embedding layer. Therefore, we allow the transformer layers to learn the relative positions of the tokens, which is more flexible and effective for capturing the contextual relationships between them.

### 3.2 Graph Building

Similar to the LUKE-Graph, we build a heterogeneous graph based on relationships stemming from the entity input. This graph accurately portrays the natural linkages between entities within a document without relying on any external knowledge graphs. In this way, we denote entity tokens (missing entity and the entities specified in the document) as the nodes of the graph. we identify three different kinds of undirected edges between them: 1) SENT-BASED edges: for node pairs that appear in the same sentences, 2) MATCH edges: for node pairs that are found in different sentences but share the same entity string, 3) PLC edges: edges between the node that corresponds to the missing entity of the question and all other nodes. An illustration of the graph creation is shown in Fig. 3.

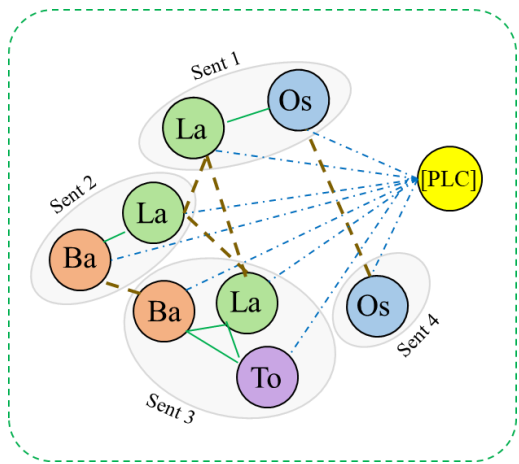

(a)          (b)

**Fig. 3.** An illustration of the graph creation

### 3.3 Relative Position Embeddings

We incorporate relative position embeddings into the attention mechanism of each transformer layer instead of absolute position in embedding layer. This approach assigns more weight to pairs of words that are closer together in the sequence and less weight to those that are farther apart. Relative position embeddings provide information on the position of tokens in the input sequence relative to each other, and they are input-length independent, making them easy to adapt to longer input sequences. To utilize these embeddings in the transformer layer, we divide the self-attention matrix into four parts: word-to-word (w2w), word-to-entity (w2e), entity-to-word (e2w), and entity-to-entity (e2e), based on the token types (word or entity). We assign a unique attention pattern to each part by using relative position labels, which are depicted in Fig. 4 using different colors. These labels are then converted into learnable vectors that modify the attention mechanism.

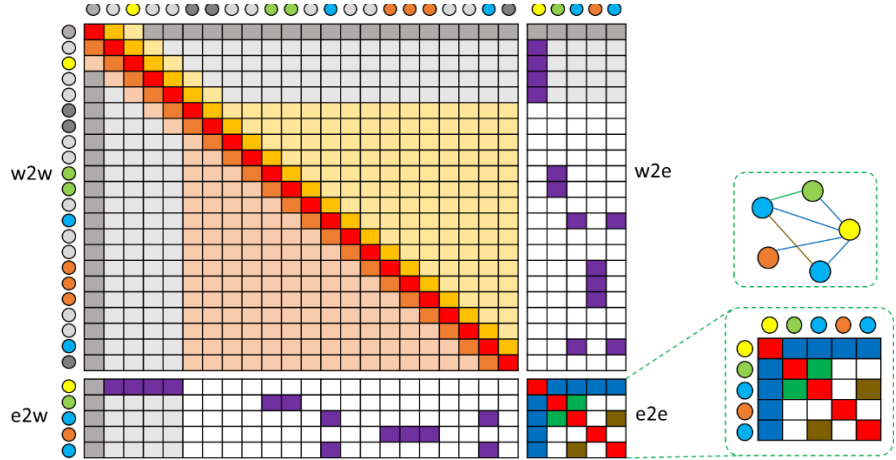

**Fig. 4.** Attention pattern used in our model. we use different colors to indicate different relative position labels.

We apply local attention, also known as sliding window attention, for w2w part. Local attention attends to a subset of positions in the input sequence and focuses on local interactions between nearby tokens, which can be computationally efficient for long sequences. The w2w part can be viewed as an undirected graph with labels on the edges connecting the word tokens. The labels depend on the relative position of the tokens, and 2k+1 (k tokens to the left, k tokens to the right and the current token) labels are defined for a given maximum distance k. Words that are further than k words to the right of word i are given the (2k+1)-th label, whereas words that are further than k words to the left of word i are given the 0-th label. Fig. 5. illustrates a schematic (Fig. 5. a) and a numerical (Fig. 5. b) lookup table of relative position labels on an input sequence of 7 words and k=2 which has a total of $(2 \times 2 + 1)$ 5 relative position embeddings ((Fig. 5. c) to be learned. Due to the fact that local attention is not adaptable enough to develop task-specific representations, similar to Longformer [10], we mark some important tokens of the w2w part ([CLS] token, question tokens) as global which attend to all positions in the input sequence. We make this attention operation symmetric; it means that a token attends to all other tokens in the sequence, and all tokens attend to it. Therefore, we assign a specific relative position label for each type of global token; one label for attending [CLS] token with other tokens and another label for attending each question token with others.

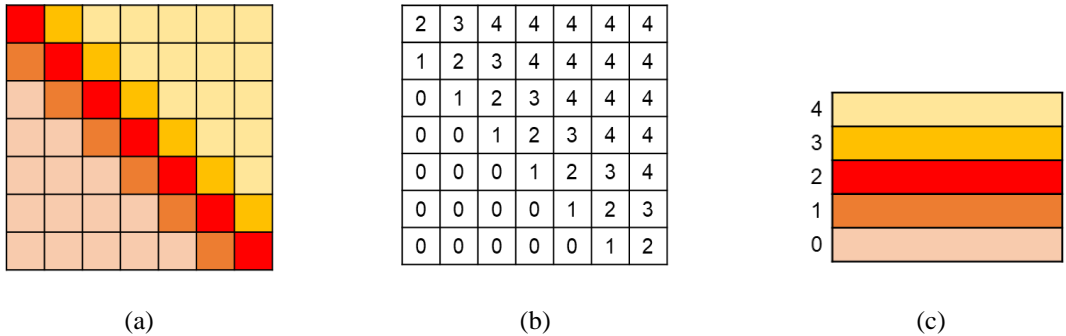

(a)  (b)  (c)

**Fig. 5.** Illustration of a schematic (Fig. 5. a) and a numerical (Fig. 5. b) lookup table of relative position labels on an input sequence of 7 words and k=2 which has a total of 5 relative position embeddings ((Fig. 5. c) to be learned.

Similar to the ETC [8], we add other relative position labels based on pairwise relationships between tokens in the w2e and the e2w parts. Therefore, one relative position label is assigned to link the entity tokens in entity input with the word tokens that belong to them in word input (their string-matched mentions), and a different label for those that do not. We also encode the created heterogeneous graph (described in Section 3.2) in e2e part. In this way, the graph is viewed with labels on the edges connecting the entity tokens, and different relative position labels are assigned for each kind of edge (SENT-BASED, MATCH, and PLC) between them.

In summary, all relative position labels used in our self-attention mechanism are listed below:
- 2k+1 labels to attend to the k left and k right neighbors of each token in the w2w part.
- One label for [CLS] token of the w2w part to attend to all positions in the input sequence.
- One label for question tokens of the w2w part to attend to all positions in the input sequence.

- One label for the w2e and e2w parts to link the entity tokens in the entity input with their string-matched mention tokens in the word input, and a different label for those that do not.
- One label for the PLC edges of the e2e part to link the missing entity (placeholder) with other entities.
- One label for the SENT-BASED edges of the e2e part to link the entities in a same sentence.
- One label for the MATCH edges of the e2e part to link the same string-matched entities in the different sentences.

### 3.4 Transformer Layers with entity-aware self-attention

The heterogeneous information regarding entities and the input sequence representations are imported to the transformer layers. In the transformer layers, we modify the entity-aware self-attention mechanism of the LUKE by adding relative position embeddings, which efficiently takes individual pairings' influence on attention towards one another into account. Furthermore, we encode a heterogeneous graph alongside each token's designated relation type as an attention pattern, which is then integrated into the entity-aware self-attention mechanism using relative position encoding.

**Self-attention:** The self-attention mechanism is a key component of the transformer architecture. In the context of the transformer, self-attention refers to a mechanism by which the model can weigh the importance of different parts of a sequence when generating its output. Similar to the LUKE [6], entity-aware self-attention is achieved through the use of token type, word or entity for computing attention score. In a self-attention layer, the input sequence (long and global input) is transformed into three vectors: the query vector, the key vector, and the value vector. These vectors are then used to compute an attention vector. In that way, an individual query vector is utilized based on the kind of tokens (word to word, word to entity, entity to word, and entity to entity). Also, the relative position embeddings are added to the query and key vectors before computing the dot product. The relative position embeddings are designed to capture the relative distance between each pair of tokens in the input sequence, allowing the model to attend to nearby tokens more strongly than those further away. To compute the attention weights, the dot product between the modified query and key vectors is divided by the square root of the dimension of the key vector, as in the standard self-attention mechanism. The resulting scores are then passed through a softmax function. Finally, the attention vector is computed as the weighted sum of the value vectors.

Formally, given an input sequence $x_1, x_2, ..., x_P$, which $x_i \in \mathbb{R}^L$ is a token representation, the attention vectors are $y_1, y_2, ..., y_P$, where $y_i \in \mathbb{R}^H$ is calculated as follows:

$$y_i = \sum_{j=1}^{P} a_{ij}(x_i W^V)$$

$$a_{ij} = \text{softmax}(\frac{x_i W^Q (x_j W^K)^T + x_i W^Q (r_{ij})^T}{\sqrt{H}})$$

$$\text{where } W^Q = \begin{cases} W^{Q_{w2w}} & \text{if } x_i \in w, x_j \in w \\ W^{Q_{w2e}} & \text{if } x_i \in w, x_j \in e \\ W^{Q_{e2w}} & \text{if } x_i \in e, x_j \in w \\ W^{Q_{e2e}} & \text{if } x_i \in e, x_j \in e \end{cases} \quad (1)$$

Where $W^Q, W^K, W^V \in \mathbb{R}^{H \times L}$ represent the learnable weight query, key, and value matrices, respectively. $r_{ij}$ denotes a learnable relative position vector assigned to the label between token i and j. There are four query matrices ($W^{Q_{w2w}}, W^{Q_{w2e}}, W^{Q_{e2w}}, W^{Q_{e2e}}$), one of which is selected based on the attention score is calculated between which the type of tokens; word ($w$) or entity ($e$). P and H refer to the input sequence length and hidden state dimension, respectively.

### 3.5 Score Accumulation

The final representation of the placeholder (PLC) is concatenated with each representation of candidate entity in the entity input. Then, a fully connected layer followed by a sigmoid function is applied on them to compute the candidate score. Finally, the model is trained using binary cross-entropy loss averaged over all candidates, and the candidate with the highest score is chosen as the best answer $e^*$:

$$e^* = \underset{e \in E}{\mathrm{argmax}}\, f_o([y_{PLC}; y_e]) \qquad (2)$$

Where $f_o(.)$ is a fully connected layer followed by a sigmoid function, $[y_{PLC}; y_e]$ is a concatenation of the PLC representation ($y_{PLC}$) and the candidate representation ($y_e$), and $E$ denotes the entity input includes set of entity candidates.

## 4 Experiments

This section is dedicated to showcasing the efficacy of our model on datasets that demand robust reasoning abilities: ReCoRD [15], emphasizing commonsense reasoning, and WikiHop [28], which requires multi-hop reasoning. We aim to assess and benchmark our model's performance against other state-of-the-art methods operating in these domains.

### 4.1 Training Configuration

The architecture of the model is based on that of the LUKE large model [6], which has 24 hidden layers, 1024 hidden dimensions ($L = 1024$), 64 attention head dimensions ($H = 64$), and 16 self-attention heads. This means that a word token embedding has 1024 dimensions, while an entity token embedding has 256 dimensions, which is then converted to 1024 dimensions using a dense layer. The input text is tokenized using RoBERTa's tokenizer [29], which has a vocabulary of 50K words, and the entity vocabulary includes 500K common entities, as well as two special entities, [MASK] and [UNK], with [UNK] being used for missing entities. The maximum sequence length ($P$) and maximum question length are set to 512 and 90, respectively, and the other hyperparameters are similar to the LUKE model that is shown in Table 1. The attention pattern is created with k=150, which controls the maximum distance between two tokens in the input sequence that can attend to each other in the w2w part. However, we adjust this value based on the length of the input sequence and the presence of longer sequences in the dataset that require a larger value. Additionally, when transferring weights from the pre-trained LUKE model, the original query matrix $W^Q$ is copied to different matrices of $W^{Q_{w2w}}, W^{Q_{w2e}}, W^{Q_{e2w}}, W^{Q_{e2e}}$ in the model. The model is trained using an Amazon EC2 p3.8xlarge instance with four GPUs, and a single model trained with 2 batch sizes and 2 epochs takes about 2.5 hours for the ReCoRD dataset and 6 hours for the WikiHop dataset. The longer training duration of WikiHop, in contrast to ReCoRD, can be attributed primarily to its larger dataset size and the introduction of an entity extraction section.

Table 1. Hyper-parameters of our model.

| Name | Value | Name | Value |
|---|---|---|---|
| Max Seq length ($P$) | 512 | number of transformer layers | 24 |
| Max question length | 90 | hidden size in transformer ($L$) | 1024 |
| Warmup ratio | 0.06 | attention head size in transformer ($H$) | 64 |
| Weight decay | 0.01 | number of self-attention heads in transformer | 16 |
| Adam β1 | 0.9 | word token embedding size | 1024 |
| Adam β2 | 0.98 | entity token embedding size | 256 |
| Adam ϵ | 1e-6 | maximum relative position distance (k) | 150 |

### 4.2 Datasets

This study utilizes the ReCoRD [15] and WikiHop [29] datasets designed to evaluate the ability of NLP models to understand and reason over complex text.

**ReCoRD:** It consists of over 120,000 cloze-style questions, which require selecting a missing entity (placeholder) to fill in the blank in a given document. For each question, the placeholder is filled with a suitable answer from all entities in the associated document. One of the unique features of ReCoRD is that it emphasizes commonsense reasoning and contextual understanding over surface-level factual recall. The dataset includes documents from a variety of domains, including science, history, and literature, and the questions require reasoning about the context to determine the correct answer. For example, a question might ask about the actions of a character in a story, or the implications of a scientific experiment. The dataset is split into 100k training, 10k development, and 10k test sets, with each set containing documents and questions from a range of difficulty levels. ReCoRD has been used as a benchmark for evaluating the performance of NLP models on tasks that require more advanced reasoning and comprehension skills.

The ReCoRD dataset uses two evaluation metrics to assess the performance of NLP models on the task of cloze-style reading comprehension. The first metric is the exact match (EM) score, which measures the percentage of questions for which the model's predicted answer exactly matches the ground-truth answer. The second metric is the F1 score, which calculates the harmonic mean of precision and recall for the predicted answer compared to the ground-truth answer. The EM metric is a strict measure of accuracy and penalizes models heavily for even minor errors in their predictions. For example, if a model predicts the correct answer but misspells it slightly, it will receive a score of 0 for that question under the EM metric. The F1 score, on the other hand, is a more forgiving measure that takes into account partial matches between the predicted and ground-truth answers.

**WikiHop:** Derived from Wikipedia, the WikiHop dataset presents a set of 51k questions, each requiring the model to synthesize information from diverse sections of the text to formulate a comprehensive answer. The dataset consists of 43k training, 5k development, and 2k test sets. It uniquely emphasizes multi-hop reasoning, mirroring real-world scenarios where effective question-answering involves the integration of knowledge from disparate sources. The question format is presented as tuples (?, property, subject), akin to [7], and has been converted to suitable format for our method. Specifically, underscores in the property have been replaced with spaces, and the question mark is substituted with a [PLC] token. For instance, the original tuple (?, record_label, get_ready) transforms into the cloze-style format "[PLC] record label get ready". Evaluation of model performance in the WikiHop dataset is based on accuracy, where the model's predicted answer is deemed correct if it precisely aligns with the ground truth answer. Notably, this accuracy metric is equivalent to the EM metric, ensuring a strict assessment of the model's ability to provide accurate and contextually relevant responses.

### 4.3 Results and Discussion

**ReCoRD:** Table 2 presents a comparison of our method with other state-of-the-art models on the ReCoRD dataset [15]. The results have been obtained from multiple sources, including the ReCoRD[1] and SuperGLUE[2] [30] leaderboards, as well as relevant literature [6,31].

Our model (GESA-500M) has been evaluated against three large language models (LLMs), namely T5-11B [25], PaLM-540B [27], and DeBERTa-1.5B [26], which outperform our model by +1.7 EM, +1.6 EM, and +2.4 EM, respectively. However, these LLMs require a significant amount of computational resources for effective training and inference, which can make them expensive and time-consuming to use. Nonetheless, comparing transformer-based models can be challenging due to variations in computing and pretraining resources. Our model differs significantly from the aforementioned LLMs, as it has only around 500 million parameters, while they have hundreds of billions of parameters. When compared to T5-Large, which has 770 million parameters, our model outperforms it by +5.4 F1 and +5.8 EM, despite being 1.5 times smaller. Our encoder-only setup is particularly suitable for tasks that require commonsense reasoning, while the decoder-encoder architecture of DeBERTa and T5 models is more versatile for various natural language processing tasks. Conversely, PaLM has a decoder-only setup and requires significant computational resources, making it less accessible to researchers or organizations without adequate computational resources, especially when deploying it on resource-constrained devices. In conclusion, the smaller number of parameters in our model provides a significant advantage over LLMs, enabling more efficient training and inference.

In comparison to other transformer-based models, some models such as BERT [1], XLNET [32], and RoBERTa [29] lack explicit knowledge of inference concepts. As a result, their representations, as shown in the BERT-based group of Table 2, fall short in their ability to support reasoning and have lower accuracy. Since entities play a crucial role in such datasets, and there is reasoning about the relationships between them, LUKE [6] outperforms RoBERTa significantly by +0.6 F1/+0.6 EM, with the only difference being the addition of entities to the inputs and the use of an entity-aware self-attention mechanism. Therefore, we have chosen LUKE as our transformer-based model. Furthermore, LUKE-Graph [7] has demonstrated that the creation of a heterogeneous graph for reasoning through the relationships of these entities can significantly enhance the model's capability to select relevant information and make better decisions. To achieve this, we have integrated the heterogeneous information with the transformer layers instead of using a separate graph module, eliminating the need for a multi-step model. As a result, our model exhibits a significant improvement over LUKE-Graph, with an increase of +0.7 F1 and +0.5 EM, while also reducing execution time. This indicates that incorporating graph information with attention is an effective approach, as demonstrated by a gain of +2.1 (F1+EM) over the LUKE.

---

[1] https://sheng-z.github.io/ReCoRD-explorer/

[2] https://super.gluebenchmark.com/tasks

Compared to the graph-based models, namely Graph-BERT [33], SKG-BERT [34], KT-NET [35], and KELM [31], our model achieves a significant improvement of +29.2 F1/+30.9 EM, +19.4 F1/+19.5 EM, +17.4 F1/+18.7 EM, and +15.5 F1/EM, respectively. While SKG-BERT, KT-NET, and KELM employ external knowledge graphs such as WordNet, ConceptNet, and NELL, our model does not rely on any additional knowledge graph. Instead, we convert document entities into a graph to capture their relationships.

**Table 2.** The results obtained from the ReCoRD dataset. It is noteworthy that all models except DeBERTa (ensemble) are based on a single model. The absence of results is denoted by a hyphen. The results marked with [±] are reported in the KELM paper, while those marked with [+] are reported in the LUKE paper. The remaining results have been obtained from various sources, including the ReCoRD and SuperGLUE leaderboards, as well as relevant literature. GESA-500M indicates our model with 500 million parameters.

| Group | Name | Dev | | Test | |
|---|---|---|---|---|---|
| | | F1 | EM | F1 | EM |
| | Human | 91.64 | 91.28 | 91.69 | 91.31 |
| BERT-based | BERT-Base [1] | - | - | 56.1 | 54.0 |
| | BERT-Large± [1] | 72.2 | 70.2 | 72.0 | 71.3 |
| | XLNet-Verifier+ [32] | 82.1 | 80.6 | 82.7 | 81.5 |
| | RoBERTa+ [29] | 89.5 | 89.0 | 90.6 | 90.0 |
| Graph-based | Graph-BERT [33] | - | - | 63.0 | 60.8 |
| | SKG-BERT± [34] | 71.6 | 70.9 | 72.8 | 72.2 |
| | KT-NET± [35] | 73.6 | 71.6 | 74.8 | 73.0 |
| | KELM± [31] | 75.6 | 75.1 | 76.7 | 76.2 |
| LLM | T5-Large [25] | - | - | 86.8 | 85.9 |
| | T5-11B [25] | 93.8 | 93.2 | 94.1 | 93.4 |
| | PaLM 540B [27] | ***94.0*** | ***94.6*** | 94.2 | 93.3 |
| | DeBERTa-1.5B [26] | 91.4 | 91.0 | ***94.5*** | ***94.1*** |
| LUKE-based | LUKE+ [6] | 91.4 | 90.8 | 91.2 | 90.6 |
| | LUKE-Graph [7] | 91.36 | 90.95 | 91.5 | 91.2 |
| | **GESA-500M** | **92.14** | **91.61** | **92.2** | **91.7** |

**WikiHop:** The results presented in Table 3 provide an extensive comparison between our model, GESA, and other cutting-edge models on the WikiHop dataset, organized into three groups: attention-pattern-based, graph-based, and LUKE-based. Particularly noteworthy is our emphasis on models within the attention-pattern-based group, demonstrating that our attention pattern surpasses in performance compared to other patterns. However, it is important to acknowledge a potential unfairness in the comparison due to the fixed input length of 4096 for all attention-pattern-based models versus the 512 input length for our model. Given WikiHop's extensive contexts, input length significantly influences model accuracy, and we plan to address this limitation by increasing the input length in the future. Despite this, GESA outperforms both Longformer [10] and ETC [8] in the test set, showcasing the efficacy of our fusion attention pattern in integrating reasoning information derived from the graph.

Furthermore, GESA exhibits significant superiority over graph-based models: Entity-GCN [36] by +15.1, BAG [37] by +13.7, HDEGraph [28] by +11.8 and Path-GCN [38] by +10.2. This notable result underscores the effectiveness of the robust transformer architecture embedded in GESA. Impressively, GESA even surpasses LUKE-based models, achieving gains of +5.6 and +1.7 over the LUKE baseline model and LUKE-Graph, respectively. This comparison underscores the efficacy of GESA in leveraging attention patterns in conjunction with the LUKE architecture, ultimately enhancing performance on intricate multi-hop reasoning tasks within the challenging WikiHop dataset.

**Table 3.** The results obtained from the WikiHop dataset. All models are based on a single model. "-" indicates missing results.

| Group | Name | Dev ACC | Test ACC |
|---|---|---|---|
| | Human | - | 74.1 |
| Attention-pattern-based | Longformer-base [10] | 75.0 | - |
| | Longformer-large [10] | 77.6 | 81.9 |
| | ETC-large [8] | ***79.8*** | 82.3 |
| | RealFormer-large [39] | 79.21 | ***84.4*** |
| Graph-based | Entity-GCN [36] | 64.8 | 67.6 |
| | BAG [37] | 66.5 | 69.0 |
| | HDEGraph [28] | 68.1 | 70.9 |
| | Path-GCN [38] | 70.8 | 72.5 |
| LUKE-based | LUKE [6] | 73.2 | 77.1 |
| | LUKE-Graph [7] | 77.8 | 81.0 |
| | **GESA** | **79.1** | **82.7** |

### 4.1 Ablation study

To examine the impact of each component on our model, we conduct multiple ablation experiments on the ReCoRD development dataset.

**Effect of using different attention patterns:** In this study, we examine the impact of different attention patterns on the performance of our model across four components: w2w, w2e, e2w, and e2e. Our findings, presented in Table 4, indicate that attention mechanisms play a crucial role in optimizing model performance. In the w2w component, we experiment by omitting attention between global [CLS] and question tokens, and only using a local attention pattern. We observed a decline in F1/EM results by 0.16/0.14, indicating the importance of attending to all tokens for some crucial tokens. Similarly, in the e2e component, we use a local attention pattern and ignore heterogenous information from the created graph in the attention mechanism, which results in a performance drop of 1.4%. This highlights the significance of integrating the graph with self-attention.

We also investigate the impact of removing relative position labels from Equation (1) in different attention components. Following LUKE [6], we use Equation (3) to compute the attention score between two tokens.

$$a_{ij} = \text{softmax}(\frac{x_i W^Q (x_j W^K)^T}{\sqrt{H}}) \tag{3}$$

When we remove all relative position labels in w2e, e2w, and e2e components, and only keep 2k+1 labels in w2w, the model's performance decreases significantly, reaching an F1 score of 90.1 and EM score of 89.65, which is lower than the accuracy of the baseline LUKE (91.4 F1/90.8 EM). The reduced performance is not only due to the removal of graph information and the relationship between related entity and word tokens, but also because the use of relative position embeddings in the w2w component limits the model to encoding only the relative positions of tokens within a certain range of each other (k). Consequently, the model may not be capable of capturing dependencies that span longer distances in the sequence.

To demonstrate the importance of relative position labels in the w2e, e2w, and e2e components, we remove 2k+1 and global labels in w2w but keep the labels in other components. Indeed, we use Equation (3) to compute the attention score between tokens in w2w. The F1/EM results degrade by 0.94/1.21, indicating that relative position labels in the three components are more critical than those in w2w.

**Table 4.** Effect of using different attention patterns on the ReCoRD dev set. w/o stands for without.

| Model | Dev Results (%) | | | |
|---|---|---|---|---|
| | F1 | Δ | EM | Δ |
| Full Model | 92.14 | - | 91.61 | - |
| Local attention pattern in w2w (w/o global tokens) | 91.98 | 0.16 | 91.47 | 0.14 |
| Local attention pattern in e2e (w/o graph attention) | 90.72 | 1.42 | 90.2 | 1.41 |
| w/o relative position labels in three parts of w2e, e2w and e2e | 90.1 | 2.04 | 89.65 | 1.96 |
| w/o relative position labels in w2w | 91.2 | 0.94 | 90.4 | 1.21 |

**Impact of relative position labels:** we investigate the effect of relative position labels by removing them individually as shown in Table 5. In our approach, we incorporate two relative position labels for global tokens in the w2w parts, one for the [CLS] token and another for question tokens. To assess the impact of each label, we individually omit them and observe a performance drop of 0.05 and 0.11 on the F1 metric, respectively. Our results reveal that the relationships of question tokens with all input tokens are more crucial than those of the [CLS] token. We also restrict the use of relative position labels between word and entity tokens in the w2e and e2w parts to only represent the missing entity (PLC) relationships with question tokens. This results in the removal of any special attention between candidate entities and their mentions, leading to a decrease in the F1 metric by 0.41. Conversely, when we remove the connection between PLC and question tokens and instead maintain the relationships between candidate entities and their mentions, the F1 metric decreases by 0.22.

To evaluate the contribution of each edge type in the constructed graph, we eliminate the corresponding labels individually and measure their impact on the model's performance. Specifically, we remove the relative position label of edges between entity pairs of the graph in the same sentences (SENT-BASED edges), the label of connections between matching entities (MATCH edges), or the label of edges between the missing entity (placeholder) with other entities (PLC edges). The performance on F1 falls off 0.43, 0.25, and 0.56, respectively. Our findings suggest that the connections between the PLC and other entities have a more substantial impact than other relationships, as removing the PLC edges leads to the most significant performance drop of 0.56 on the F1 metric. We also investigate the effect of using a single label for all SENT-BASED, MATCH, and PLC edges instead of three labels, which results in a decrease of F1 by 0.27.

**Table 5.** Ablation results of relative position labels on the ReCoRD dev set. w/o stands for without.

| Attention part | Model | Dev Results (%) | | | |
|---|---|---|---|---|---|
| | | F1 | Δ | EM | Δ |
| | Full Model | 92.14 | - | 91.61 | - |
| w2w | w/o [CLS] global label | 92.09 | 0.05 | 91.57 | 0.04 |
| | w/o global label of question tokens | 92.03 | 0.11 | 91.51 | 0.1 |
| w2e and e2w | w/o label between candidate entities and their mentions | 91.73 | 0.41 | 91.14 | 0.47 |
| | w/o label between PLC and question tokens | 91.92 | 0.22 | 91.42 | 0.19 |
| e2e | w/o label of SENT-BASED edges | 91.71 | 0.43 | 91.22 | 0.39 |
| | w/o label of MATCH edges | 91.89 | 0.25 | 91.4 | 0.21 |
| | w/o label of PLC edges | 91.58 | 0.56 | 90.98 | 0.63 |
| | w one label for all edges | 91.87 | 0.27 | 91.32 | 0.29 |

## 5 Future Works

Our model faces a clear limitation with its input length of 512, particularly struggling with longer documents like those found in the WikiHop Dataset. To tackle this issue, we aim to improve efficiency by extending the input length through the integration of the Longformer approach. Furthermore, in future work, we plan to utilize the capabilities of LLMs with prompt learning methods [40,41] to generate meaningful rationales and explanations for each dataset sample, enhancing the overall representation of input data.

## 6 Conclusion

In this paper, we proposed a graph-enhanced self-attention approach that extends upon the pre-trained multi-layer transformer of the LUKE model. Our approach incorporates several modifications to address the limitations of complex reasoning tasks in machine reading comprehension. Specifically, we introduced a unique attention pattern that includes global-local attention for word tokens, graph attention for entity tokens, and attention for related word and entity tokens to enhance performance. The attention pattern is integrated into the self-attention mechanism of the transformer using relative position encoding. Furthermore, a heterogeneous graph is constructed based on relationships from the entity input, which exhibits strong attention towards tokens connected in the graph in the attention mechanism. Our proposed model also considers the relationship type between each entity token and word token, resulting in more efficient attention between them if they are related. Experimental results indicate that our model outperforms both the LUKE-Graph and the baseline LUKE model on the ReCoRD dataset with commonsense reasoning.

In conclusion, our proposed model effectively integrates heterogeneous graph information into the entity-aware self-attention mechanism of LUKE using relative position labels, which has the potential to improve the performance of various natural language processing tasks that require the handling of graph structures.